\documentclass{article}

\usepackage[preprint]{explainableAI}

\usepackage[utf8]{inputenc} % allow utf-8 input
\usepackage[T1]{fontenc}    % use 8-bit T1 fonts
\usepackage{hyperref}       % hyperlinks
\usepackage{url}            % simple URL typesetting
\usepackage{booktabs}       % professional-quality tables
\usepackage{amsfonts}       % blackboard math symbols
\usepackage{nicefrac}       % compact symbols for 1/2, etc.
\usepackage{microtype}      % microtypography
\usepackage{xcolor}         % colors
\usepackage{graphicx}
\usepackage{amsmath}

\title{Active Globally Explainable Learning for Medical Images via Class Association Embedding and Cyclic Adversarial Generation}

\author{%
	Ruitao Xie\textsuperscript{1} \quad Jingbang Chen \quad  Limai Jiang \quad  Rui Xiao \quad Yi Pan \quad Yunpeng Cai\textsuperscript{1}
\\
~\\
	Shenzhen Institute of Advanced Technology, Chinese Academy of Sciences\\
~\\
	Shenzhen, 518055, China\\
~\\
	\textsuperscript{1}\{rt.xie, yp.cai\}@siat.ac.cn \\
}

\begin{document}

	\maketitle

	\begin{abstract}
		Explainability poses a major challenge to artificial intelligence (AI) techniques. Current studies on explainable AI (XAI) lack the efficiency of extracting global knowledge about the learning task, thus suffer deficiencies such as imprecise saliency, context-aware absence and vague meaning. In this paper, we propose the class association embedding (CAE) approach to address these issues. We employ an encoder-decoder architecture to embed sample features and separate them into class-related and individual-related style vectors simultaneously. Recombining the individual-style code of a given sample with the class-style code of another leads to a synthetic sample with preserved individual characters but changed class assignment, following a cyclic adversarial learning strategy. A multi-class adversarial discriminator is applied for stimulating generation of real-looking samples with expected class assignments. A multi-criteria training scheme with pair-wise style shuffling is adopted to efficiently learn the separation of the class-style and individual-style subspaces by feeding randomly paired sample from different classes. Class association embedding distills the global class-related features of all instances into a unified domain with well separation between classes. The transition rules between different classes can be then extracted and further employed to individual instances. We then propose an active XAI framework which manipulates the class-style vector of a certain sample along guided paths towards the counter-classes, resulting in a series of counter-example synthetic samples with identical individual characters. Comparing these counterfactual samples with the original ones provides a global, intuitive illustration to the nature of the classification tasks. We adopt the framework on medical image classification tasks, which show that more precise saliency maps with powerful context-aware representation can be achieved compared with existing methods. Moreover, the disease pathology can be directly visualized via traversing the paths in the class-style space.
	\end{abstract}

	\section{Introduction}
	The “black-box” problem, or the lack of explainable inferences, poses a major and sustaining challenge for a wide range of machine learning models, especially deep neural networks [1]. Unexplainable models suffer risks of systematic defects such as short-cut learning [2] and vulnerability to adversarial attacks [3], which is unbearable for critical applications such as medical diagnosis [4]. Explainable models are favorited in many decision-making scenarios because they not only increase reliability and user recognition, but also can aid knowledge discovery. Hence, in recent years there’s an increasing interest in deriving explanations from machine learning models, or developing intrinsic interpretable models [5], which both fall in the category of explainable AI (XAI).

	Several approaches have been developed aiming to produce explanation from certain machine learning models. Typical approaches include gradient and perturbation-based attribution methods, rule-based methods and knowledge-distillation based methods [6-9]. Moreover, these methods can be divided into two classes. Global explanations strive to extract sample-independent knowledge reflecting general classification rules, while local explanations identify key factors associated with the classification of individual samples. The vast majority of existing XAI methods are local ones, including some gradient-based methods, e.g., saliency maps [10], Grad-CAM [11] and Grad-CAM ++ [12], some perturbation-based methods, e.g., LIME [13], Occlusion [14], SFL [15] and others [16-19]. These methods are easy to implement and to interpret. However, they merely provide rough association weights between features and class labels, without further knowledge on the classification rules. Moreover, due to the pervasive nonlinearity and the complex interactions between features and context, the resulted explanations are often biased [20], unstable [21], inaccurate and often misleading [22, 23]. Some researches attempt to adopt transparent models (such as trees or equations [24-26]) or knowledge distillation [6-9] to achieve global XAI. However, simple and transparent interpretable models are with low prediction accuracy in most cases, thus are weak in resolving the behavior of the original model.
	
	Counterfactual generation [27, 28] is a perturbation-based approach which works on context-aware, semantic-level features (such as super-pixels or word terms) rather than raw inputs, and employs generative adversarial networks for producing synthetic samples that respects the data distribution, surpassing heuristic in-filling ones in this way. A recent approach [29, 30] generate exemplars and counter-exemplars through an adversarial autoencoder [31], which map the data into a latent space through an encoder, and apply a random perturbation on the latent vector to decode real-looking synthetic samples knowing as exemplars and counter-exemplars. These works suggested an approach to explain machine learning models by observing how the prediction results shift between semantically modified synthetic samples. However, current methods can only generate valid samples by minor perturbing of one instance, thus are not capable of achieving global explanation.

	To achieve accurate XAI, a model that explain a system of quantitative rules learnt from the entire dataset but can vividly apply to each individual sample would be more favorited than existing approaches[32, 33], where only local responses or rough global trends are given. In quite a wide range of applications, machine learning models handle objects belonging to the same domain but partitioned into various classes (or grades), many of which are also naturally transitable across classes or grades while keeping individual characters. For example, a patient developing tumors from health status would have pathogenic changes on his CT images but with other features retained on the images. Intuitively, if we can model the transition rules and freely manipulate one sample (e.g., image) to make it transit across different classes (or grades), we obtain an explainable model for the target discriminative task as well as the behavior of the classifiers. However, one major problem would be that it is practically impossible to obtain sufficient number of paired samples for supervised learning. For example, in medical AI it is rarely possible to keep the complete record of patients from health to disease that span a long period. Even if the data is complete, one single patient may not traverse all possible pathogenic paths. Synthetic data is important to tackle this difficulty.
	
	Following the idea, we propose a new approach for XAI called active explainable learning, where explanations are achieved by creating transit paths in a learnt low-dimensional manifold that allow converting a sample into an emulated one with the looking of different classes but preserving its background characters, through which the domain knowledge underlying the discriminative rules can be explored and visualized. We encode the image samples into a vector space which is separated into two sub-spaces, a style subspace encoding the class-related features, and a background subspace encoding sample-wise features that are conditionally independent to the classes. Combining the background code of one sample and the style code of another sample would produce a real-looking synthetic sample which has analogue individual features as the former one but would be (most likely) classified the same as the latter one by a certain classifier. To achieve this, we adopt a symmetric cyclic general adversarial network (GAN) with a multi-class discriminator that identifies real and fake images as well as predicting class labels at the same time. A multi-criteria training scheme with pair-wise style shuffling is applied to enable the separation of the style and background subspaces. By mapping the class-related features of all samples into the unified, class-separable code space, the global class transition rules can be then explored and visually explained by freely manipulating the class-individual code combination and synthesizing desired new data. We summarize our contributions as follows:
	
	\begin{itemize}
		\item We propose a novel XAI technique called class association embedding (CAE) which learns a separated class-individual pair of embedded representations for all data, and develop an active global explainable learning framework based on the above technique. This approach is distinct from existing XAI approaches in the following scopes: (1) Unlike existing XAI methods which provide only local perturbation analysis or rough global trends, we illustrate the full transition paths from one class to another that can assign to each individual sample, which is more powerful in depicting context-aware, nonlinear and non-stationary rules, while at the same time provide accurate and detailed explanation to each instance. (2) Unlike most current XAI methods that work on existing input features such as pixels or super-pixels, we compress the feature set in a low-dimensional embedded space, which more efficiently represents context-aware knowledge and leads to precise and detailed explanations. (3) We provide an active form of exploring the embedded knowledge by manipulating a certain sample for altering its class assignments but keeping its individual characters, thus depicting the global rules on individual data in an easy-to-understand manner.
		\item We propose a cyclic generative adversarial network to implement the aforementioned model, which features in a pair-wise code shuffling training scheme to efficiently learn the separation of the two subspaces. Unlike previous style-GAN methods which treat image style as an individual sample feature that try to mimic, we work on exploring the relationship between styles of different samples using a paired supervision manner, and thus reaches a low-dimensional, class-continuous and class-separated feature space for depicting classification rules.
		\item Medical artificial intelligent is one field that explainability is badly demanded to aid doctor decision while existing XAI approaches are weak in providing useful insights. We adopt the method on medical image classification tasks, which show that our method not only produces more precise saliency maps compared with existing XAIs, but also preserves the locality of sample features in the manifold and help with obtaining classification rules in line with existing medical knowledge, thus is efficient.
	\end{itemize}

	\section{Methods}
	\subsection{Class Association Embedding using a Cyclic Generative Adversarial Network}
	Explainable learning works on a black-box model $b(x)=y$ with a series of input-output pairs $\{x_n,y_n\}_{(n=1..N)}$ and $y \in \mathbb{C}$, by generating a explanation ${\Psi}(x,y)$ for the black-box model. The form of $\Psi$ is versatile but in many existing methods $\Psi (x_n,y_n)$ includes a saliency map indicating the importance of variables in $x_n$ that lead to the classification decision $y_n$. $\Psi$ can be either model-dependent or model-agnostic. Our method falls in the latter category in the sense that it only requires $\{x_n,y_n\}_{(n=1..N)}$ as inputs without knowing the details of $b$.
	
\begin{figure}
		\centering
		\includegraphics[scale=0.4]{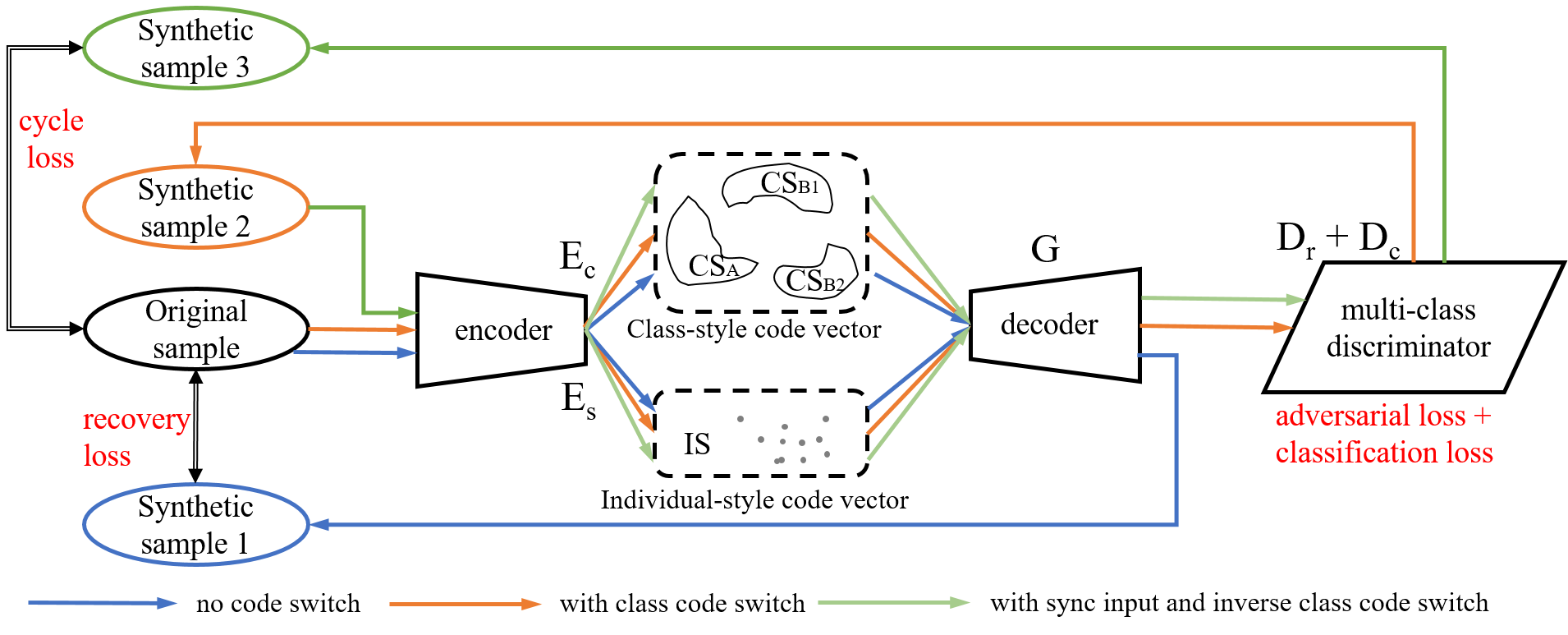}
		\caption{The Overall framework of Class Association Embedding. Synthetic sample $1$ (generated from the original sample) and Synthetic sample $3$ (generated from synthetic sample $2$)  are expected to resemble the original sample; while synthetic sample $2$ (generated from the original sample) is expected to inherit the individual style of the original sample (class $A$), but with the classification features of class  $B1$ or $B2$ (so that it would be classified as $B1$ or $B2$ using a third-party classifier).    
		}
		\label{overall-framework}
	\end{figure}

	The proposed class association embedding framework is shown in Figure \ref{overall-framework}, which includes a encoder $E$, a decoder $G$ and a multi-class adversarial discriminator $D$. The encoder consists of two modules, $E_c$ which encodes class-related style codes ($CS$, $c$ for simplicity), and $E_s$ which encodes individual-related style codes ($IS$, $s$ for simplicity), respectively. The decoder takes both the class-style and individual-style code vectors as input to generate a new synthetic sample. The adversarial discriminator concurrently generates two outputs which attempts to discriminate real and synthetic samples ($Dr$) while at the same time deterimining the proper class assignments ($Dc$).

	To handle medical image data used in this paper, we implement the framework with CNN-based deep neural networks. We equip the individual-style encoder with three convolution layers and six residual blocks (each has two convolution layers) and the class-style encoder with six convolution layers. The residual blocks in the decoder are equipped using Adaptive Instance Normalization (AdaIN) [34] layers whose parameters are obtained by a multilayer perceptron (MLP). The discriminator was designed with four convolution layers, two residual blocks each with two convolution layers and two fully connected layers. 
	%The detailed architecture is depicted in the appendix.
	
 	\subsection{Learning the Separation of Code Spaces with Random Shuffling Paired Training}
	The idea of cyclic adversarial generation with style codes has been previously proposed (e.g., [35-39]). However, the requirements of style-controlled generation in XAI are completely different from previous image generation schemes. For previous image generation tasks, “style” is a subjective visual effect defined on individual images and algorithms mainly focus on emulating the style of a given image, without considering the possible relationships between different styles. Hence previous algorithms do not strive to gather style codes into classes or clusters. However, for XAI application proposed here, in order to describe the transition rules between classes, the style-code space need to gather styles of the same class together and separate different classes, and show a smooth, low-dimensional manifold that can generalize outside training samples, which cannot be achieved by existing style-GAN methods. 

	In order to learn the efficient representation of the class-related style in a unified space, we propose a training method called random shuffling paired training. Specifically, we randomly select two samples from different classes from the training set, which are fed into the same encoder to generate individual-style codes and class-style codes respectively. A code-swapping scheme is performed so that the recombination codes can be decoded into two synthetic samples, with switched class assignments to each other. A second-round combination is performed by switching the class-style code back and the new synthetic samples are expected to recover their original class assignments. 
	The training details are given in Appendix A.4. 
	After training with a large combination of sample pairs, the class-style and individual-style subspaces can be separated successfully.
	
	\subsection{Design of Loss Function}
	Loss functions are designed to efficiently learn the class-association embedding vectors and achieve the cyclic adversarial learing goals. Let $A:(x_A,y_A)$ be the sample to be credited and $B:(x_B,y_B)$ be its paired sample (whose loss should be credited separately). Let $E_{c}$, $E_{s}$ and $G$ be the mapping functions of the encoders and decoders described above. The first part of the loss function is the reconstuction loss. The basic reconstruction loss, shown in equation \ref{con:loss1}, represents the error of the encode-decode loop without any code switch. Moreover, we argue that modifying the class-style code of a sample should not affect its individual-style code, and vice versa. Hence, we encode the synthetic sample and compare its class-style and individual-style codes with their original counter parts, resulting in the class-wise and sample-wise reconstruction loss denoted in equations \ref{con:loss2} and \ref{con:loss3}, which help with improving the consistency of the code space when training.
 
	\begin{equation}
		\mathcal{L}_{recon}^{x_{A}}=\mathbb{E}_{x_{A} \sim p(x_{A})}\left[ {\parallel G(E_{c}(x_{A}),E_{s}(x_{A}))-x_{A} \parallel}_{1} \right]
		\label{con:loss1}
	\end{equation}
	\begin{equation}
		\mathcal{L}_{recon}^{c_{A}}=\mathbb{E}_{c_{A} \sim p(c_{A}), s_{B} \sim q(s_{B})} \left[ {\parallel E_{c}(G(c_{A},s_{B}))-c_{A} \parallel}_{1} \right]
		\label{con:loss2}
	\end{equation}
	\begin{equation}
		\mathcal{L}_{recon}^{s_{A}}=\mathbb{E}_{c_{B} \sim p(c_{B}), s_{A} \sim q(s_{A})} \left[ {\parallel E_{s}(G(c_{B},s_{A}))-s_{A} \parallel}_{1} \right]
		\label{con:loss3}
	\end{equation}

	The second part of the loss functin is the cyclic loss as shown in equation \ref{con:loss4}, which calculates the error of recovering the original sample after the two-round encode-decode cycle with two style switchs. The third and fourth parts are the adversarial and classification loss functions, shown in equations \ref{con:loss5} and \ref{con:loss6}. Where $D_{r}$ represents the output probability of the sample being real-looking ($1$ refers to real, $0$ fake), and $D_{c}$ the probability of assigning the correct class ($y_A$).
	\begin{equation}
		\mathcal{L}_{cyc}^{x_{A}}=\mathbb{E}_{x_{A} \sim p(x_{A})}\left[ {\parallel G(c_{A}, E_{s}(G(c_{B}, s_{A}) ))-x_{A} \parallel}_{1} \right]
		\label{con:loss4}
	\end{equation}
	\begin{equation}
		\mathcal{L}_{adv1}^{A2B}=\mathbb{E}_{c_{B} \sim p(c_{B}), s_{A} \sim q(s_{A})} \left[ -log(\frac{exp(D_{r}(G(c_{B},s_{A}))[1])}{exp(D_{r}(G(c_{B},s_{A}))[0])+exp(D_{r}(G(c_{B},s_{A}))[1])})  \right]
		\label{con:loss5}
	\end{equation}
	\begin{equation}
		\mathcal{L}_{cla1}^{A2B}=\mathbb{E}_{c_{B} \sim p(c_{B}), s_{A} \sim q(s_{A})} \left[ -log(\frac{exp(D_{c}(G(c_{B},s_{A}))[y_{B}])}{\sum_{j \in \mathbb{C}} exp(D_{c}(G(c_{B},s_{A}))[j])})  \right]
		\label{con:loss6}
	\end{equation}
%	The calculation of these loss functions during shuffling paired training is also given in Supplemental Figure \ref{training}. 
	Combinging the above loss functions (with $\lambda_{1}$, $\lambda_{2}$, $\lambda_{3}$, $\lambda_{4}$, $\lambda_{5}$, $\lambda_{6}$ as weights), we use the object function in equation \ref{con:loss7} to train and update the encoder and decoder.
	
	\begin{equation}
		\begin{split}
			\mathcal{L}(E, G)=\lambda_{1}(\mathcal{L}_{recon}^{x_{A}}+\mathcal{L}_{recon}^{x_{B}})+\lambda_{2}(\mathcal{L}_{recon}^{c_{A}}+\mathcal{L}_{recon}^{c_{B}})
			+\lambda_{3}(\mathcal{L}_{recon}^{s_{A}}+\mathcal{L}_{recon}^{s_{B}})\\+\lambda_{4}(\mathcal{L}_{cyc}^{A}+\mathcal{L}_{cyc}^{B})+\lambda_{5}(\mathcal{L}_{adv1}^{A2B}+\mathcal{L}_{adv1}^{B2A})+\lambda_{6}(\mathcal{L}_{cla1}^{A2B}+\mathcal{L}_{cla1}^{B2A})
			\label{con:loss7}
		\end{split}
	\end{equation}
	
	On the other hand, the parameters in the discriminator $D$ are updated based on equation \ref{con:loss8}, \ref{con:loss9} and \ref{con:loss10}, which are calculated on each pair of samples together. 
	
	\begin{equation}
		\begin{split}
			\mathcal{L}_{adv2}^{A2B}=\mathbb{E}_{c_{B} \sim p(c_{B}), s_{A} \sim q(s_{A})} \left[ -log(\frac{exp(D_{r}(G(c_{B},s_{A}))[0])}{exp(D_{r}(G(c_{B},s_{A}))[0])+exp(D_{r}(G(c_{B},s_{A}))[1])}) \right] \\
			+\mathbb{E}_{x_{B} \sim p(x_{B})} \left[-log(\frac{exp(D_{r}(x_{B})[1])}{exp(D_{r}(x_{B})[0])+exp(D_{r}(x_{B})[1])}) \right]
			\label{con:loss8}
		\end{split}
	\end{equation}
	\begin{equation}
		\mathcal{L}_{cla2}^{A}=\mathbb{E}_{x_{A} \sim p(x_{A})} \left[ -log(\frac{exp(D_{c}(x_{A})[y_{A}])}{\sum_{j \in \mathbb{C}} exp(D_{c}(x_{A})[j])})  \right]
		\label{con:loss9}
	\end{equation}
	\begin{equation}
		\mathcal{L}(D)=\varphi_{1}(\mathcal{L}_{adv2}^{A2B}+\mathcal{L}_{adv2}^{B2A})+\varphi_{2}(\mathcal{L}_{cla2}^{A}+\mathcal{L}_{cla2}^{B})
		\label{con:loss10}
	\end{equation}
	For each iteration, we firstly optimize the discriminator while fixing the encoder and decoder, and then update the other two closely.

	\section{Experiments}
	\subsection{Data and Implementation Details}
\begin{figure}
		\centering
		\includegraphics[scale=0.3]{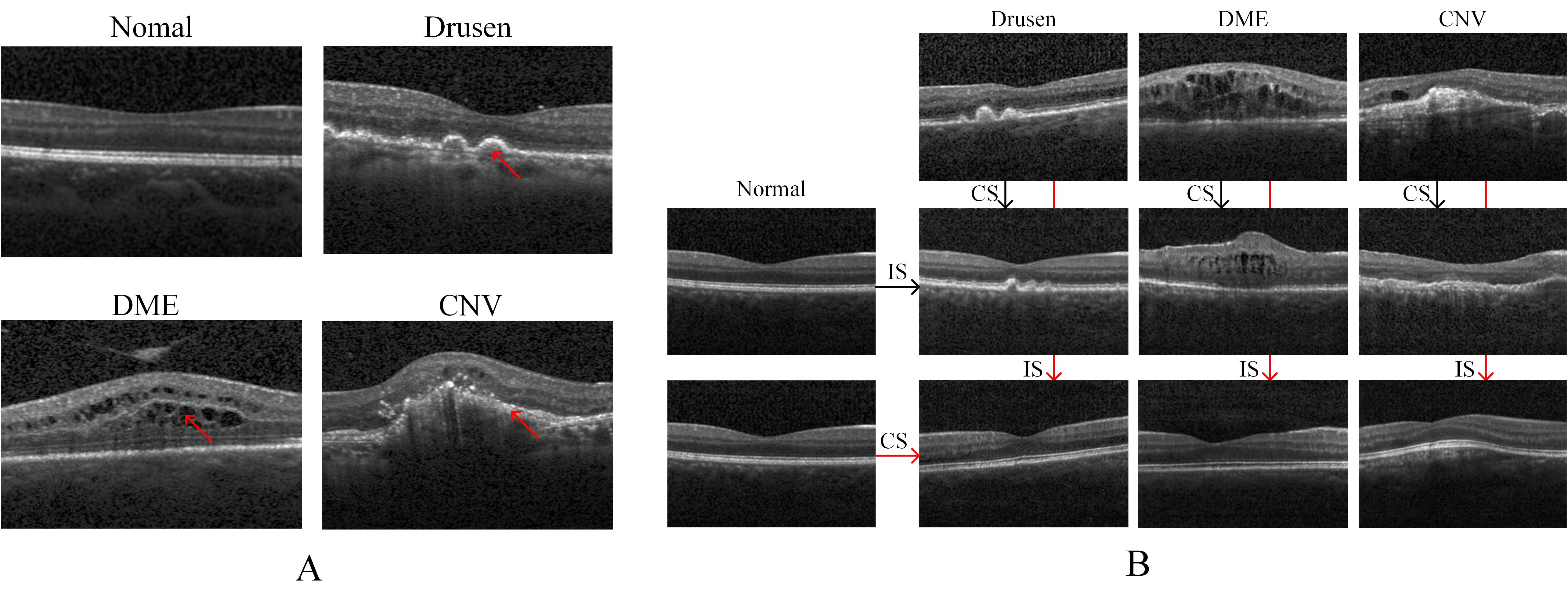}
		\caption{($A$) Four pathological classes of OCT images: normal, drusen (pointed by the arrow), diabetic macular edema (DME, with retinal-thickening-associated intraretinal fluid as pointed by the arrow), choroidal neovascularization (CNV, with neovascular membrane as pointed by the arrow). ($B$): synthetic images from normal to abnormal (middle row) and from abnormal to normal (bottom row).}
		\label{oct-cases-gen}
	\end{figure}

	We use a large retinal Optical Coherence Tomography (OCT) image dataset published by Kermany et al [40] in 2018 for verifying our proposed approach. Unlike most existing XAI methods that use the same samples to build and validate explanations, we separate the data into training and testing datasets to show that our explanation is replicable across data. The dataset contains 108312 OCT images for training, which are from 4686 patients, and consists of 51140 images with normal retina, 37206 with choroidal neovascularization (CNV), 11349 with diabetic macular edema (DME) and 8617 with drusen, while the test set contains 1000 OCT images (250 in each class). In order to make training data more balanced, we randomly select 8000 images from the original training set in each class and form new one in our experiment. We show several cases in Figure \ref{oct-cases-gen}.A. To better verify our algorithm, we also take Pathologic Myopia Challenge (PALM) dataset [41] for saliency map generation tasks, which includes 400 color fundus images labeled (213 are normal and the remaining ones are abnormal, 320 images are randomly selected as our training set and the rest part is used for testing).
	
	In our experiments, input images are center-cropped to be square and resized as 256×256. Random horizontal flip with probability of 0.5 for input images is taken for data enhancement when training. We use Adam optimizer for parameters updating with both initial learning rate and weight decay set to 1e-4. Empirically, we set the weights $\lambda_{1}$, $\lambda_{2}$, $\lambda_{3}$, $\lambda_{4}$, $\lambda_{5}$, $\lambda_{6}$ in equation 7 as $10$, $1$, $1$, $10$, $1$, $1$, and the weights $\varphi_{1}$ $\varphi_{2}$ in equation 10 as $1$, $2$ separately in our experiments.

	\subsection{Successful Feature Transmission and Class Re-assignments}
	The basic idea of our approach is to generate real-looking samples with arbitrary swapped class assignments. To validate this, we perform experiments on the retinal OCT dataset and show some synthetic cases in Figure \ref{oct-cases-gen}.B. We can see that the generated images are real-looking and possess similar class-related features of the paired images whose class-style codes are employed, while also retaining the identity-related background characteristics (contour, structure, etc.) of the original images. In order to objectively verify the class-swapping implementation performance of image generation, we introduce an external classifier using resnet50 [42] (trained and tested on the same set and achieves 94.5\% accuracy) for testing synthetic samples. In the test set, 92.64\% of the normal-to-disease transition achieved identical classification results (including identical disease subtypes) with the external classifier while for the disease-to-normal transition the conincidence rate is 93.73\%. These results prove the success of class-style transition using our proposed method. Furthermore, in case the external classifier used is the target model for our explanation, the inconsistly classified samples can be iteratively used as new training inputs so that the explanation can be improved.

	\subsection{Class Association Embedding Learns Pathological(Class)-Related Sample Manifolds}
\begin{figure}
		\centering
		\includegraphics[scale=0.3]{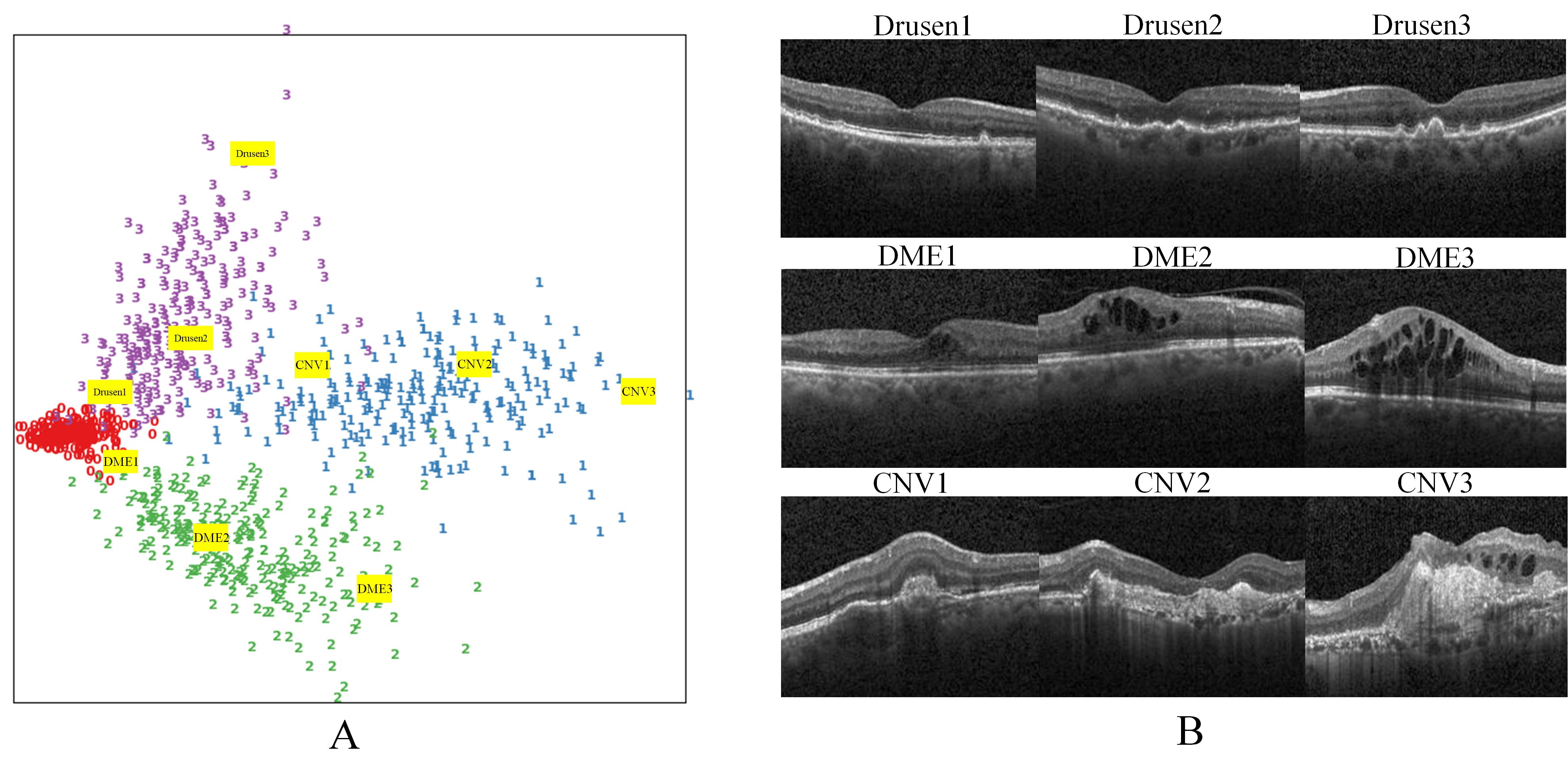}
		\caption{($A$): The class-style space for 4 OCT classes ($0$: normal, $1$: CNV, $2$: DME, $3$: Drusen). ($B$): Detailed orignal images at the sampled paths.}
		\label{style-distance-cases}
	\end{figure}

	We extract the class-style codes (each with 8 dimensions) of the OCT images in the test set encoded by the trained model and perform a principal component analysis (PCA), displayed in Figure \ref{style-distance-cases}.A. We see that samples from images with different pathological classes are separated apart in the test set as expected, which proves that the learned class-associated embedding can be replicable to external data. Furthermore, we observe that class $CNV$ and class $Drusen$ are close to each other, and class $CNV$ is farther away from class $Normal$ than class $Drusen$. Also $Drusen$ samples are postioned adjuncent to the path from $Normal$ to $CNV$. These are in accordance with existing medical knowledge: The development of $Drusen$ may transit into $CNV$. Furthermore, we randomly select three set of samples along linear paths starting from the center of the normal class, as marked in the figure. The corresponding original images are presented in Figure \ref{style-distance-cases}.B. We see a continuous growing trend of lensions along the path with similar shapes, despite their heterogeneous source. This means that CAE is locality-preserving in the sense that it can cluster samples of similar class-style into adjacent positions. Hence the embedded data manifold can be used to explore explainable rules that coincide with domain knowledge.

	\subsection{Active and Global Explanation of Classification Rules via Continuous and Guided Image Generation}
	 
\begin{figure}
		\centering
		\includegraphics[scale=0.21]{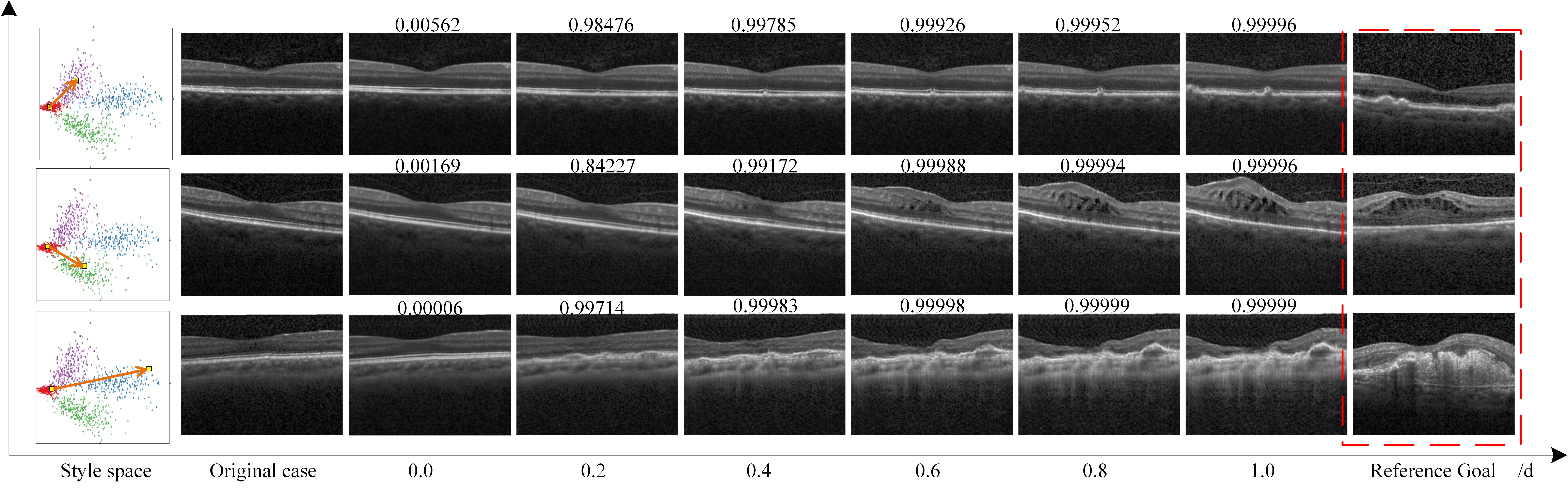}
		\caption{Generated images based on dragged class-style codes (indicated by the brown rows) with different relative distances from the origin.}
		\label{drag-gen}
	\end{figure}

	In many applications such as clincal diagnosis, class-related transition is continuous and dynamic but the acquisition of images is costly, making it infeasible to observe a set of continuous state transformation on the same object/person. The inter-object differences frustrate not only the classification quality, but also the understanding and visualizing of the classification rules. With class association embedding, we are able to eliminate inter-object differences and depict the comprehensive knowledge learned from the entire dataset on one single sample, which provides a more efficient means of explanation. To demostrate this, in the above OCT image dataset we randomly select two samples of different classes (one normal and one abnormal) and create a linear path between them in the class-style space. The individual-style code of the normal sample is used combining with different class-style code sampled evenly on the path, resulting in a series of synthetic images as depicted in Figure \ref{drag-gen}. We observe that as the class-style codes move from the normal group towards the abnormal one, the pathological features of the corresponding synthetic images evloves with similar lesion character, which vividly uncovers the underlying classification/pathological rules. By setting up various paths in the class-style space, a set of different classification rules with clear medical meaning can be convieniently explored.
	
	\subsection{Improving the Accuracy of Saliency Maps with Class Association Embedding}
	
\begin{figure}
		\centering
		\includegraphics[scale=0.295]{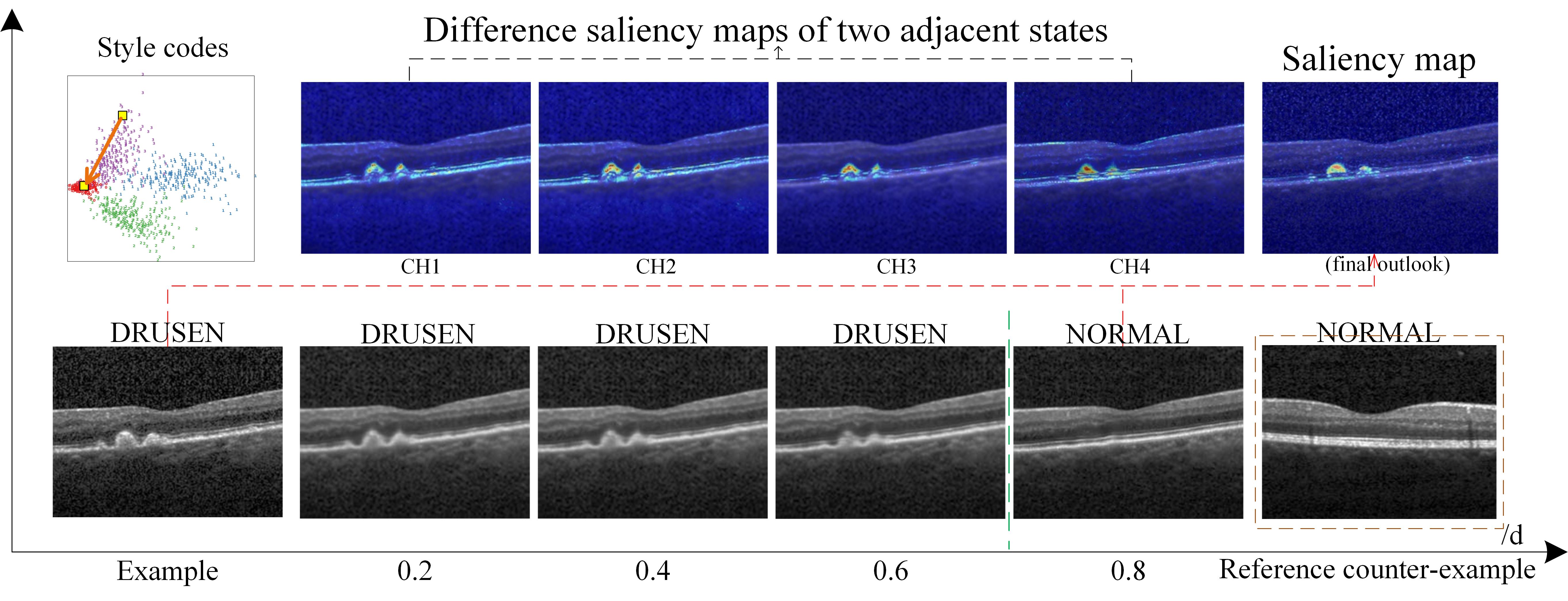}
		\caption{Example showing the generation of difference maps and saliency maps based on serial counter-example generation using CAE.}
		\label{explain1}
	\end{figure}

	Generating a saliency map is often the most favorited part of explanation, especially in image recognition tasks. With the manipulatable feature of CAE we are able to generate a series of intermediate samples and contrasting them to reach a more precise and detailed location of ROIs (the regions of interest), which surpasses existing methods where only local perturbation or pair-wise contrast can be adopted. Furthermore, separating the individual-style and class-style spaces also make CAE less prone to sample imbalances and noises, which also improve the quality of the saliency maps. In this section we illustrate a procedure to generate high-accuracy saliency maps using CAE. Specifically, for each instance to be explained, a cross-class path is created as mentioned in the above subsection with synthetic samples generated as counter-examples using the class-style code sampled in the path and the individual-style code of the instance, as is shown in Figure \ref{explain1}. The classification probabilities for each synthetic example can also be calculated using either the inner or the third-party discriminators. The images are aligned and the frame-to-frame differences are calculated, resulting in a series of differential maps $CH_{n}$. The saliency map can be obtained by summing up all $CH_{n}$ with weights propotional to their classification probability changes, or in many cases more simply by contrasting the destinational image to the original one if the feasible transition path is linear. The destination point in the style space can be either manually chosen, or automatically determined by choosing the first sampled point that completely flip the class assgnment along the given path.
	
\begin{figure}
		\centering
		\includegraphics[scale=0.195]{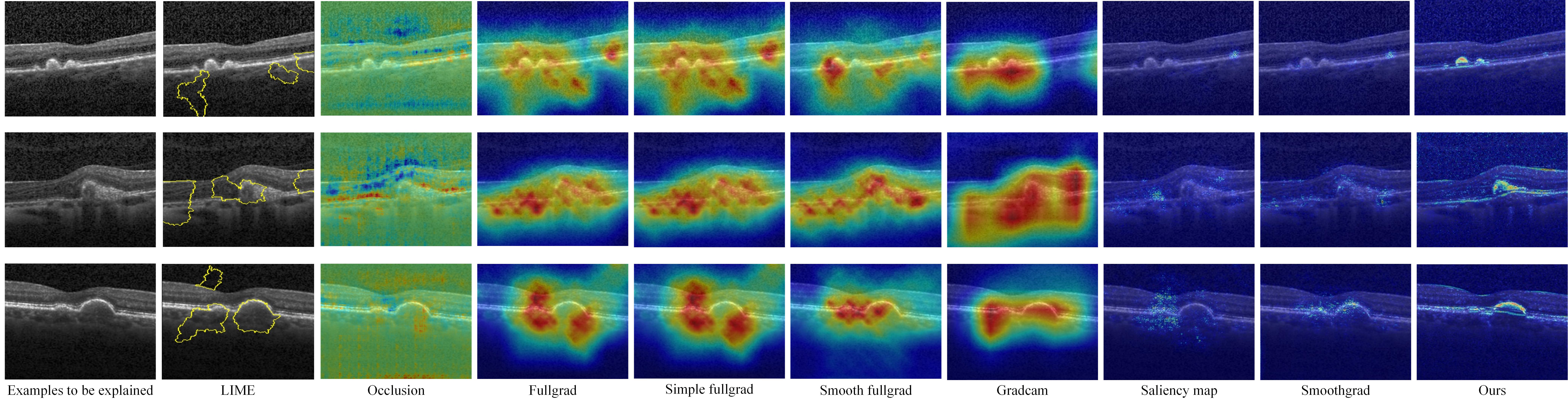}
		\caption{Compared results with other XAI methods, including $LIME$ [13], $Occlusion$ [14], $Fullgrad$ [43], $Simple$ $fullgrad$ [43], $Smooth$ $fullgrad$ [43], $Grad-CAM$ [11], $Saliency$ $map$ [10], $Smoothgrad$ [44].}
		\label{explain2}
	\end{figure}

	To illustrate the advantage of our approach, we compare our approach with several state-of-the-art XAI methods in Figure \ref{explain2} using the same external classifier as the target black-box model. We can see the regions of interest (ROIs) we get are more accurate, more fine-grained and with clearer contours. Comparing with existing methods which are mostly passive and perturbation-based, our algorithm is more in line with the global pathological rules, more sophisticated but computationally efficient with guided transition paths. The guided manipulation in our method following the global class-related manifolds enables global and context-aware knowledge visualization. Moreover, for medical images where the inter-class differences can be subtle and the dataset can be small-sized and noisy, many existing methods (e.g., LIME) suffer from data biases or noises and generate false ROIs as is shown in the figure. Our method avoids this problem by removing the divergenced background features which lead to more task-specific explanations.

\begin{figure}
		\centering
		\includegraphics[scale=0.195]{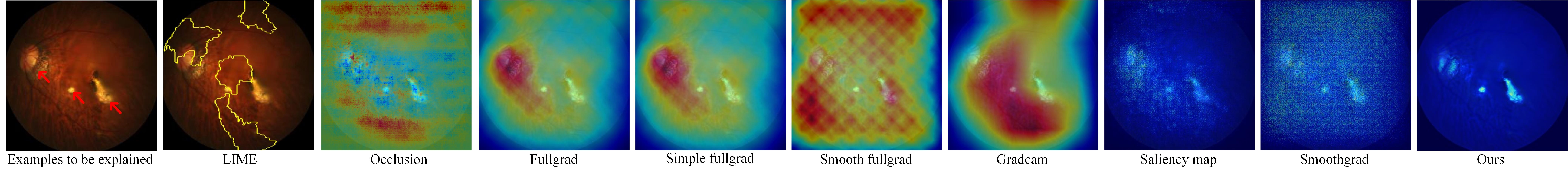}
		\caption{The fundus case with its generated saliency maps using our proposed method and other compared ones from PALM dataset [41]. The regions where the red rows point are dieases-related.}
		\label{other-data-explain}
	\end{figure}

	To verify the generalization ability of our proposed framework, we conduct experiments for explaining classification tasks on other dataset and show the results of case explained with ours and other methods in the Figure \ref{other-data-explain}, showing that our proposed approach outperforms others as is discussed.
	
	\section{Conclusion}
	In summary, we propose an active global XAI framework based on the proposed technique of class association embedding for effectively addressing the explainability challenge in medical AI. Global and class-related knowledge of all samples are distilled into a unified, class-discriminative style space with powerful representation of pathological relations, where transition rules between classes can be further revealed and visually explained by actively manipulating the embedded class-style codes and synthesizing a series of new identity-preserving samples with continuously changing class-related features along the guided paths, which greatly improves the interpretability and operability of exploring and describing context-aware pathogenetic rules from samples. These cannot be achieved by current XAI methods and are greatly favored by medical researchers. More detailed and accurate saliency maps are obtained using this framework compared with other existing XAI methods.
	
	Additionally, class association embedding preserves the following desirable abilities (validations given in Appendix): 1) The class-style space is continuous, meaning that a convex recombination of multiple style-codes from the same class would generate a sample with the same class assignment; 2) The class-style is semantically pervasive, meaning that applying the same code on different background images would create similar style-transferring effects; 3) The computation burden of generating saliency maps is far less than perturbation or counterfactual methods, because there’s no need to generate a lot of locally perturbed samples; The framework can be interactive by iteratively generate new samples with different style-codes and query them in the black-box model to reach in-depth understanding of the model. Hence, we believe that class association embedding would be a potentially powerful tool for representing and exploring domain knowledge in AI models. The major limitation of our work is that whether CAE can fully represents the model behaviors and how to maximize its represented behaviors remains unknown, which is crucial in avoiding decision risks caused by AI models and will be our future directions.

	\section*{References}
	
	\medskip

	{
		\small

		[1] Singh A, Sengupta S, Lakshminarayanan V. Explainable deep learning models in medical image analysis[J]. Journal of Imaging, 2020, 6(6): 52.
		
		[2] DeGrave A J, Janizek J D, Lee S I. AI for radiographic COVID-19 detection selects shortcuts over signal[J]. Nature Machine Intelligence, 2021, 3(7): 610-619.
		
		[3] Xu M, Zhang T, Li Z, et al. Towards evaluating the robustness of deep diagnostic models by adversarial attack[J]. Medical Image Analysis, 2021, 69: 101977.
		
		[4] Karar M E, Hemdan E E D, Shouman M A. Cascaded deep learning classifiers for computer-aided diagnosis of COVID-19 and pneumonia diseases in X-ray scans[J]. Complex \& Intelligent Systems, 2021, 7(1): 235-247.
		
		[5] Pintelas E, Livieris I E, Pintelas P. A grey-box ensemble model exploiting black-box accuracy and white-box intrinsic interpretability[J]. Algorithms, 2020, 13(1): 17.
		
		[6] Tan S, Caruana R, Hooker G, et al. Distill-and-compare: Auditing black-box models using transparent model distillation[C]//Proceedings of the 2018 AAAI/ACM Conference on AI, Ethics, and Society. 2018: 303-310.
		
		[7] Frosst N, Hinton G. Distilling a neural network into a soft decision tree[J]. arXiv preprint arXiv:1711.09784, 2017.
		
		[8] Che Z, Purushotham S, Khemani R, et al. Distilling knowledge from deep networks with applications to healthcare domain[J]. arXiv preprint arXiv:1512.03542, 2015.
		
		[9] Liu X, Wang X, Matwin S. Improving the interpretability of deep neural networks with knowledge distillation[C]//2018 IEEE International Conference on Data Mining Workshops (ICDMW). IEEE, 2018: 905-912.
		
		[10] Simonyan K, Vedaldi A, Zisserman A. Deep inside convolutional networks: Visualising image classification models and saliency maps[J]. arXiv preprint arXiv:1312.6034, 2013.
		
		[11] Selvaraju R R, Cogswell M, Das A, et al. Grad-cam: Visual explanations from deep networks via gradient-based localization[C]//Proceedings of the IEEE international conference on computer vision. 2017: 618-626.
		
		[12] Chattopadhay A, Sarkar A, Howlader P, et al. Grad-cam++: Generalized gradient-based visual explanations for deep convolutional networks[C]//2018 IEEE winter conference on applications of computer vision (WACV). IEEE, 2018: 839-847.
		
		[13] Ribeiro M T, Singh S, Guestrin C. " Why should i trust you?" Explaining the predictions of any classifier[C]//Proceedings of the 22nd ACM SIGKDD international conference on knowledge discovery and data mining. 2016: 1135-1144.
		
		[14] Zeiler M D, Fergus R. Visualizing and understanding convolutional networks[C]//European conference on computer vision. Springer, Cham, 2014: 818-833.
		
		[15] Sun Y, Chockler H, Huang X, et al. Explaining image classifiers using statistical fault localization[C]//European Conference on Computer Vision. Springer, Cham, 2020: 391-406.
		
		[16] Bach S, Binder A, Montavon G, et al. On pixel-wise explanations for non-linear classifier decisions by layer-wise relevance propagation[J]. PloS one, 2015, 10(7): e0130140.
		
		[17]  Zintgraf L M, Cohen T S, Adel T, et al. Visualizing deep neural network decisions: Prediction difference analysis[J]. arXiv preprint arXiv:1702.04595, 2017.
		
		[18] Koh P W, Liang P. Understanding black-box predictions via influence functions[C]//International conference on machine learning. PMLR, 2017: 1885-1894.
		
		[19] Fong R C, Vedaldi A. Interpretable explanations of black boxes by meaningful perturbation[C]//Proceedings of the IEEE international conference on computer vision. 2017: 3429-3437.
		
		[20] Shrikumar A, Greenside P, Kundaje A. Learning important features through propagating activation differences[C]//International conference on machine learning. PMLR, 2017: 3145-3153.
		
		[21] Adebayo J, Gilmer J, Muelly M, et al. Sanity checks for saliency maps[J]. Advances in neural information processing systems, 2018, 31.
		
		[22] Chen L, Cruz A, Ramsey S, et al. Hidden bias in the DUD-E dataset leads to misleading performance of deep learning in structure-based virtual screening[J]. PloS one, 2019, 14(8): e0220113.
		
		[23] Kindermans P J, Hooker S, Adebayo J, et al. The (un) reliability of saliency methods[M]//Explainable AI: Interpreting, Explaining and Visualizing Deep Learning. Springer, Cham, 2019: 267-280.
		
		[24] Letham B, Rudin C, McCormick T H, et al. Interpretable classifiers using rules and bayesian analysis: Building a better stroke prediction model[J]. The Annals of Applied Statistics, 2015, 9(3): 1350-1371.
		
		[25] Giudici P, Raffinetti E. Shapley-Lorenz eXplainable artificial intelligence[J]. Expert Systems with Applications, 2021, 167: 114104.
		
		[26] Lundberg S M, Erion G, Chen H, et al. From local explanations to global understanding with explainable AI for trees[J]. Nature machine intelligence, 2020, 2(1): 56-67.
		
		[27] Yu J, Lin Z, Yang J, et al. Generative image inpainting with contextual attention[C]//Proceedings of the IEEE conference on computer vision and pattern recognition. 2018: 5505-5514.
		
		[28] Chang C H, Creager E, Goldenberg A, et al. Explaining image classifiers by counterfactual generation[J]. arXiv preprint arXiv:1807.08024, 2018.
		
		[29] Guidotti R, Monreale A, Matwin S, et al. Explaining image classifiers generating exemplars and counter-exemplars from latent representations[C]//Proceedings of the AAAI Conference on Artificial Intelligence. 2020, 34(09): 13665-13668.
		
		[30] Guidotti R, Monreale A, Giannotti F, et al. Factual and counterfactual explanations for black box decision making[J]. IEEE Intelligent Systems, 2019, 34(6): 14-23.
		
		[31] Makhzani A, Shlens J, Jaitly N, et al. Adversarial autoencoders[J]. arXiv preprint arXiv:1511.05644, 2015.
		
		[32] Kim B, Wattenberg M, Gilmer J, et al. Interpretability beyond feature attribution: Quantitative testing with concept activation vectors (tcav)[C]//International conference on machine learning. PMLR, 2018: 2668-2677.
		
		[33] Oramas Mogrovejo J A, Wang K, Tuytelaars T. Visual explanation by interpretation: Improving visual feedback capabilities of deep neural networks[C]//https://iclr. cc/Conferences/2019/AcceptedPapersInitial. openReview, 2019.
		
		[34] Huang X, Belongie S. Arbitrary style transfer in real-time with adaptive instance normalization[C]//Proceedings of the IEEE international conference on computer vision. 2017: 1501-1510.
		
		[35] Zhu J Y, Park T, Isola P, et al. Unpaired image-to-image translation using cycle-consistent adversarial networks[C]//Proceedings of the IEEE international conference on computer vision. 2017: 2223-2232.
		
		[36] Choi Y, Choi M, Kim M, et al. Stargan: Unified generative adversarial networks for multi-domain image-to-image translation[C]//Proceedings of the IEEE conference on computer vision and pattern recognition. 2018: 8789-8797.
		
		[37] Karras T, Laine S, Aila T. A style-based generator architecture for generative adversarial networks[C]//Proceedings of the IEEE/CVF conference on computer vision and pattern recognition. 2019: 4401-4410.
		
		[38] Xie X, Chen J, Li Y, et al. Self-supervised cyclegan for object-preserving image-to-image domain adaptation[C]//European Conference on Computer Vision. Springer, Cham, 2020: 498-513.
		
		[39] Park T, Zhu J Y, Wang O, et al. Swapping autoencoder for deep image manipulation[J]. Advances in Neural Information Processing Systems, 2020, 33: 7198-7211.
		
		[40] Kermany D S, Goldbaum M, Cai W, et al. Identifying medical diagnoses and treatable diseases by image-based deep learning[J]. Cell, 2018, 172(5): 1122-1131. e9.
		
		[41] Huazhu Fu, Fei Li, José Ignacio Orlando, Hrvoje Bogunovic, Xu Sun, Jingan Liao, Yanwu Xu, Shaochong Zhang, Xiulan Zhang, "PALM: PAthoLogic Myopia Challenge", IEEE Dataport, 2019. [Online]. Available: http://dx.doi.org/10.21227/55pk-8z03. Accessed: Jul. 08, 2019.
		 
		[42] He K, Zhang X, Ren S, et al. Deep residual learning for image recognition[C]//Proceedings of the IEEE conference on computer vision and pattern recognition. 2016: 770-778.
		
		[43] Srinivas S, Fleuret F. Full-gradient representation for neural network visualization[J]. Advances in neural information processing systems, 2019, 32.
		
		[44] Smilkov D, Thorat N, Kim B, et al. Smoothgrad: removing noise by adding noise[J]. arXiv preprint arXiv:1706.03825, 2017.

	}

	\appendix
	~\\
	~\\
	~\\
	\section{Appendix}

	\subsection{Continuity of Style Domain}
	The class-style code described in Sect. 2 are only extracted from sample data which are discrete and limited in number. To further analyze whether the embedded rules can be generalized to the out-of-sample regions, we adopt SMOTE (Synthetic Minority Over-Sampling Technique) to generate a set of 2000 resampled new codes (which were the convex combination of existing class-style sample codes and distributed on the surface of the manifold contour) for each category in the OCT test data set, then combining the new class-style codes with a individual-style code to generate new synthetic samples. We calculate the ratios of the new generated samples with successful class assignments, which are 93.44\% (Normal), 97.20\% (CNV), 94.11\% (DME), 97.58\% (Drusen) respectively. The wrongly assigned data may be caused by imprecise manifold contour caused by limited data. Therefore, we are confident that our class-style domain is continuous and separable, which means most of the class-style codes falling into the same subspace are with the same class association and the synthetic samples based them can be attributed to the same category.
	
	\subsection{Semantic Pervasiveness of the Class-style Codes}
\begin{figure}[!h]
		\centering
		\includegraphics[scale=0.265]{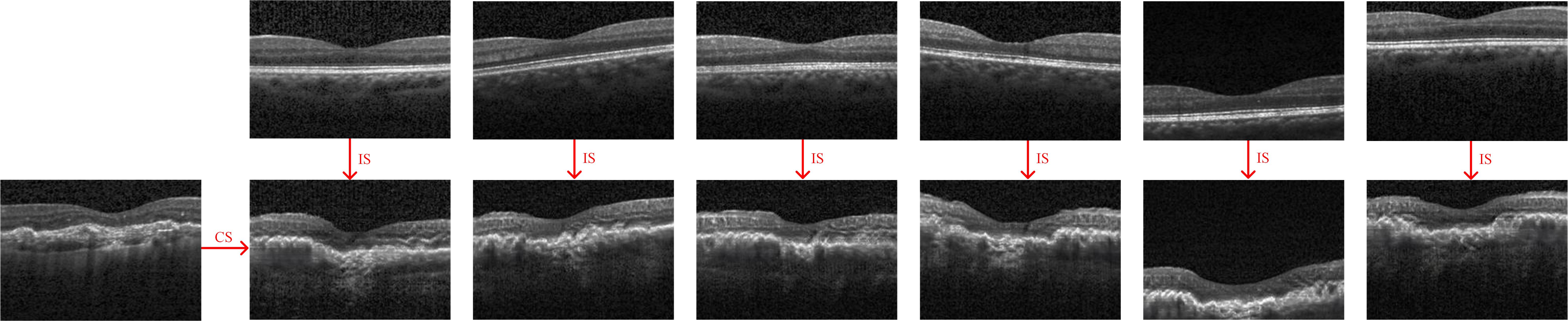}
		\caption{Cases that combine the class-style code with different individual-style codes for synthesizing new samples.}
		\label{semantic-all}
	\end{figure}

	Class association embedding is the key of our proposed XAI algorithm, and we are supposed to ensure that the pathological classes of generative images are related to the combined class-style codes and regardless of the individual-style ones. We analyze the semantic pervasiveness of the class-style codes. Specifically, we extract the class-style codes of the OCT test images which are correctly predicted by the classifier as mentioned in section 3.2, and each code is randomly combined with 10 individual-style codes from other test images predicted accurately by the classifier for generating new images. We then employ the same classifier for class prediction of these synthetic samples. The ratio of the generated cases with successful class assignments reaches 94.31\%. To illustrate semantic pervasiveness of class-style codes more intuitively, we show some cases that combine the class-style code with different individual-style codes for synthesizing new samples in Figure \ref{semantic-all}. We can see that similar class-related features are well preserved when different background images are combined in these cases, further demonstrating the powerful semantic pervasiveness of the class-style codes.
	
	\subsection{Low Computation Burden for Saliency Maps Generation}
	With guided path in the class-style space, we can actively synthesize counter examples for comparing and generating saliency maps with low computation burden. Using PyTorch software libarary V1.7.1 on a single NVIDIA TITAN Xp 12GB and Intel(R) Xeon(R) CPU E5-2650 v4@ 2.20GHz, we compute the cost of the running time that one case is explained based our proposed approach and compared with other perturbation algorithms as presented in Table \ref{compared-results-computation}. It can be seen that our XAI method far outperformed others, which further proves that with guided strategy, meaningful and efficient generations based global perturbance that are content-aware and in line with pathological rules could be implemented.   
	\begin{table}
		\caption{Compared results with other perturbation methods in terms of the computation burden.}
		\label{compared-results-computation}
		\centering
		\begin{tabular}{llllll}
			\toprule
			Methods     & LIME [13]    & Occlusion [14] & SFL [15] &Ours \\
			\midrule
			Computation cost (s) &18.90 &135.82 &147.00 & 0.15   \\
			\bottomrule
		\end{tabular}
	\end{table}

	\subsection{Detailed Illustration for Random Shuffling Paired Training}
\begin{figure}[!h]
	\centering
	\includegraphics[scale=0.32]{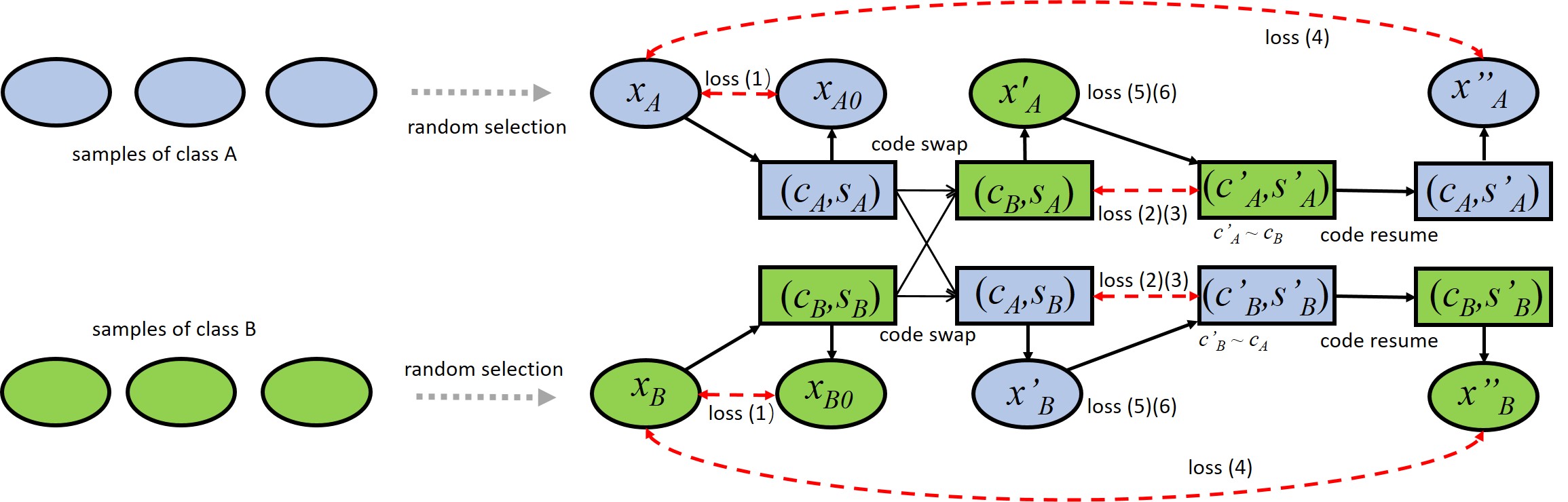}
	\caption{The Schema for Random Shuffling Paired Training.}
	\label{training}
	\end{figure}

	The detailed schema for random shuffling paired training is presented in Figure \ref{training}. The numbers in parentheses show the portion of loss functions given by the corresponding equation in section 2.3. As is pointed out, existing cyclic adversarial learning schemes do not fit in our architecture. In order to learn the efficient representation of the class-related style in a unified domain, we randomly select two images from different classes from the training set for pairing. Without loss of generality, for two paired samples $x_A$ and $x_B$, the two classes can be denoted by $y_A$ and $y_B$, respectively. Multi-class tasks can be also learned in a 1-vs-1 manner. The paired samples ($x_A$ and $x_B$) are fed into the same encoder to generate individual-style codes ($s_A$ and $s_B$) and class-style codes ($c_A$ and $c_B$) respectively. Decoding $(c_A,s_A)$ and $(c_B,s_B)$ resembles $x_A$ and $x_B$. A code-swith scheme is performed so that the combination $(c_B,s_A)$ and $(c_A,s_B)$ lead to two synthetic samples $x_A'$ and $x_B'$ with switched class assignments $y_A'=y_B$ and $y_B'=y_A$. Both samples are then re-encoded as $(c_A',s_A')$ and $(c_B',s_B')$ with $c_A' \sim c_B$ and $c_B' \sim c_A$. A second-round combination is peformed and thus $(c_A, s_A')$ and $(c_B, s_B')$ are decoded into $x_A'' \sim x_A$ and $x_B'' \sim x_B$, respectively. During shuffling random paired training process, Without loss of generality, an sample can be randomly (or enumeratedly when the data size is small) paired with many samples of different classes to generate a large number of class-individual style combinations. After enough iterations of training, class-style and individual-style subspaces are separated successfully.

\end{document}